%% file: main.tex
\documentclass[runningheads]{llncs}

\usepackage{eccv}

\usepackage{eccvabbrv}

\usepackage{graphicx}
\usepackage{booktabs}
\input{paper/preamble}
\usepackage[accsupp]{axessibility}  % Improves PDF readability for those with disabilities.

\usepackage{hyperref}

\usepackage{orcidlink}

\begin{document}

\title{Open-Set Biometrics: Beyond Good Closed-Set Models}


\author{Yiyang Su\orcidlink{0009-0002-4652-6828} \and
Minchul Kim\orcidlink{0000-0003-1449-9124} \and
Feng Liu\orcidlink{0000-0003-2103-4659} \and
Anil Jain \and
Xiaoming Liu\orcidlink{0000-0003-3215-8753}}

\authorrunning{Y.~Su \etal}

\institute{Michigan State University, East Lansing, MI 48824, USA \\
\email{\{suyiyan1, kimminc2, liufeng6, jain, liuxm\}@msu.edu}}

\maketitle

\input{paper/0_abstract}
\input{paper/1_intro}
\input{paper/2_related_work}
\input{paper/3_method}
\input{paper/4_experiments}
\input{paper/5_conclusions}

\bibliographystyle{splncs04}
\bibliography{paper/refs}

\newpage\setcounter{equation}{0}
\setcounter{figure}{0}
\setcounter{table}{0}
\setcounter{page}{1}
\setcounter{section}{0}

\begin{center}
\textbf{\Large Open-Set Biometrics: Beyond Good Closed-Set Models}\\
\vspace{2mm}
\textbf{\large Supplementary Material}\\
\end{center}

\input{paper/9_supp}

\end{document}

%% file: paper/preamble.tex
\usepackage[dvipsnames]{xcolor}

\usepackage{nicematrix}
\usepackage{pifont} % for checkmark
\usepackage{siunitx} % for displaying large numbers
\newcommand{\smallpm}{{\scriptstyle \pm}}
\usepackage{wrapfig}

%% file: paper/0_abstract.tex
\begin{abstract}
   Biometric recognition has primarily addressed closed-set identification, assuming all probe subjects are in the gallery. 
   However, most practical applications involve open-set biometrics, where probe subjects may or may not be present in the gallery. This poses distinct challenges in effectively distinguishing individuals in the gallery while minimizing false detections. 
   While it is commonly believed that powerful biometric models can excel in both closed- and open-set scenarios, existing loss functions are inconsistent with open-set evaluation. They treat genuine (mated) and imposter (non-mated) similarity scores symmetrically and neglect the relative magnitudes of imposter scores. To address these issues, we simulate open-set evaluation using minibatches during training and introduce novel loss functions: (1) the \emph{identification-detection} loss optimized for open-set performance under selective thresholds and (2) \emph{relative threshold minimization} to reduce the maximum negative score for each probe. Across diverse biometric tasks, including face recognition, gait recognition, and person re-identification, our experiments demonstrate the effectiveness of the proposed loss functions, significantly enhancing open-set performance while positively impacting closed-set performance.
   Our code and models are available \href{https://github.com/prevso1088/open-set-biometrics}{here}.
  \keywords{Open-set biometrics \and Face recognition \and Gait recognition \and Person reID}
\end{abstract}

%% file: paper/1_intro.tex
\section{Introduction}
\label{sec:intro}

\begin{figure}[tb]
    \centering
    \includegraphics[width=\linewidth, trim={13 0 15 0},clip]{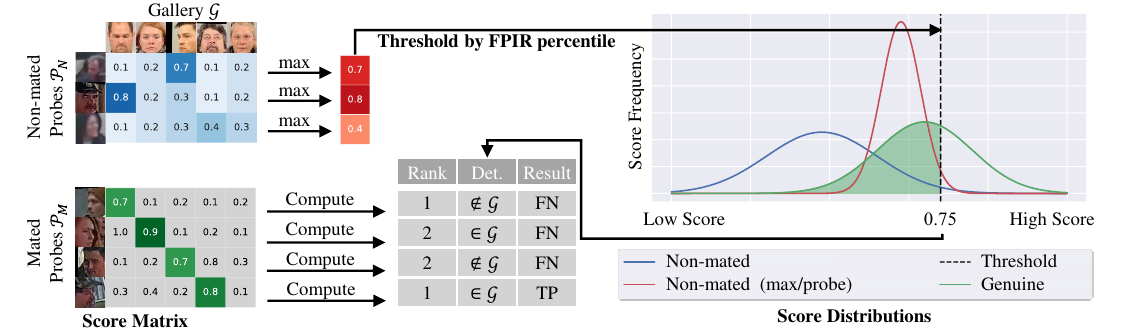}
    \caption{A toy example for FNIR@FPIR calculation. The score matrix and distribution between non-mated probes and all gallery subjects are shown in blue. The maximum score for each non-mated probe is shown in red. A percentile of their distribution determines the threshold. 
    A user-defined percentile (\eg, 33\%) sets the threshold to be 0.75 as 1 out of 3 (the 2\textsuperscript{nd}) non-mated probes is incorrectly classified as ``in the gallery''.
    The genuine scores of mated probes are shown in green. A mated probe is an FN if either of the conditions is met: (1) in detection (Det.), it is incorrectly classified as ``not in the gallery'' (\(\mathcal{G}\)) as the genuine score is below the threshold (the 1\textsuperscript{st} mated probe); (2) its genuine score is not within a specified rank \(R\), \eg, 1, in the gallery (the 2\textsuperscript{nd} mated probe).
    Hence, FNIR@$33\%$FPIR is $75\%$.
    }
    \label{fig:fnir_fpir}
\end{figure}
Biometrics, the study of unique physiological or behavioral characteristics for human identification, plays a pivotal role in applications such as access control, surveillance, and forensic analysis~\cite{ross2019some}. 
Each biometric task, from face recognition, gait recognition to person re-identification (reID), uses features tailored to its specific requirements~\cite{gait_survey, zahra2023person, liu2024farsight}.

In real-world applications, after test subjects are enrolled in the gallery, biometric systems inevitably encounter \emph{non-mated} subjects, those not present in the gallery. 
Thus, it becomes imperative for such systems to demonstrate robust performance in the \emph{open-set} framework with non-mated probes~\cite{open-set_lfw, ijb-s, open_set_fingerprint, frvt}. 
The \emph{closed-set} framework assumes all probe subjects are mated and predominantly focuses on \emph{identification}, associating probes with the closest match in the gallery. 
However, the open-set framework introduces an additional challenge: \emph{detection}, determining whether a probe subject is mated. 
For example, in airport surveillance, faces in CCTV videos are compared against a criminal watch list. 
The goal is to reduce the misidentification (false positives) of non-mated subjects such as airport staff and innocent passengers (improving detection), while simultaneously ensuring accurate recognition of known criminals (enhancing identification).
For this reason, open-set biometrics is a harder and more general task than closed-set.

The biometric community has predominantly centered on the closed-set framework.
While some face recognition benchmarks have open-set evaluation protocols~\cite{ijb-s, open-set_lfw, gunther2017unconstrained}, they are primarily used for evaluation rather than for developing tailored strategies to address the open-set challenges.
In gait recognition and person reID, most benchmarks only provide official closed-set protocols, lacking open-set ones. 
To address this gap, we take the initiative to introduce open-set benchmarks and propose strategies to train models specifically for improving open-set performances while maintaining closed-set performance.

The false negative identification rate at a given false positive identification rate (FNIR@FPIR) is a \textit{de facto} open-set metric~\cite{frvt}.
We first compute the similarity score matrix between non-mated probes and the gallery, from which we extract the maximum score for each non-mated probe across the gallery, forming the maximum-per-probe non-mated score distribution. This distribution represents the likelihood that a non-mated probe has a match in the gallery. A threshold is set based on a user-defined percentile (\eg, $1\%$) of this distribution. Mated probes with scores below this threshold are counted as false negatives (FNs), as they are incorrectly considered not to be in the gallery. Similar to the closed-set scenario, a mated probe is also an FN if its gallery match doesn't rank among the gallery set within a user-specified rank \(R\). FNIR is the proportion of mated probes classified as FNs. \cref{fig:fnir_fpir} depicts its calculation pipeline.

Unlike closed-set or verification metrics, FNIR@FPIR introduces unique challenges. 
It employs a threshold based on maximum scores for each non-mated probe, creating an asymmetry between genuine and imposter scores.
SOTA biometric recognition methods achieve strong performance in closed-set scenarios leveraging triplet loss~\cite{triplet_person_reid} and/or softmax cross-entropy based losses~\cite{adaface, arcface, cosface, magface, curricularface}, both of which excel in reducing intra-person variance and increasing inter-person variance. However, they do not take into account the fact that the non-mated score distribution determines the threshold for the false negative. 

In other words, conventional loss functions, while effective in closed-set scenarios, exhibit limitations in addressing the nuances of the FNIR@FPIR metric, which necessitate a handling of the maximum of imposter scores. These high-scoring non-mated probes disproportionately influence the threshold setting for false negatives, yet conventional loss functions often treat them equivalently to other scores. As a result, they face challenges in open-set scenarios. 
GaitBase~\cite{opengait}, a SOTA gait recognition model, can only achieve a $61\%$ FNIR@1\%FPIR on the training set and $86\%$ on the test set. 
This challenge extends to face recognition and person reID.

To address this challenge, we propose to create simulated open-set evaluation episodes during training, whose score distributions serve as proxies to those of the test set.
Specifically, we introduce an \textit{identification-detection} loss tailored to optimize the two core elements in open-set biometrics, \ie, detection and identification.
For detection, the model is trained to accurately pinpoint the exact probes whose mate is present in the gallery set; for identification, the model learns to establish associations between the gallery and probe samples of mated subjects.
By integrating the aspects of detection and identification within training batches, we improve the model's efficacy in both domains, thus yielding enhanced testing performance. 
Further, we minimize similarity scores between non-mated subjects and the gallery, enabling a more comprehensive detection of mated subjects (see \cref{fig:flow_overview}).

Empirically, we apply our proposed loss function to three biometric modalities, face, gait, and person reID, and show performance improvement in all tasks. While our focus is open-set biometrics, we also observe performance gain in closed-set search due to improved feature representations.

To summarize, our contributions include:
\begin{itemize}
    \item[\ding{51}] We show that the existing loss functions, \eg, triplet loss and softmax loss, do not treat genuine and imposter training samples in accordance with open-set evaluation. This inconsistency sheds light on why the open-set challenges have not been adequately addressed.
    \item[\ding{51}] We propose a novel loss function that leverages open-set mini-batches during training to optimize identification and detection performance and minimize open-set detection thresholds.
    \item[\ding{51}] We experimentally demonstrate that our proposed loss function can effectively improve both closed- and open-set performance across diverse biometric modalities, including face, gait, and person reID.
\end{itemize}

%% file: paper/2_related_work.tex
\section{Related Work}

\subsubsection{Open-Set Biometrics.}
While various biometric recognition tasks, such as face recognition~\cite{adaface, arcface, cosface, magface, curricularface, sphereface, cfsm, kim2024keypoint, qin2023swinface, huang2020improving, kim2022cluster}, gait recognition~\cite{gait3d, gaitedge, gaitgci, gaitgl, gaitnetv1, gaitnetv2, gaitpart, gaitset, metagait, opengait, lidargait, swingait, lagrangegait, mmgaitformer, cstl, 3d_local, ccpg, DyGait, hstl, danet, mtsgait, ye2024biggait}, and person reID~\cite{agrl, tclnet, psta, cal, ccpg, 3d_invar_reid, ni2023part, lee2023camera, mekhazni2023camera, somers2023body, chen2023towards,Liu_2024_CVPR}, have witnessed significant progress, the majority of their focus remains on closed-set scenarios. Although EVM has been applied to open-set face recognition~\cite{open-set_lfw}, EVM takes extracted features, instead of images/videos, as input. Our approach improves feature representation and EVM improves score calculation. While some face recognition datasets provide opens-set evaluation protocols~\cite{ijb-s}, open-set performance is often perceived as a consequence rather than a direct objective in advancing recognition models. 
Further, even the most recent gait recognition~\cite{ye2024biggait} and person reID~\cite{Liu_2024_CVPR} models are exclusively evaluated under closed-set settings due to the absence of open-set evaluation protocols.

\begin{figure}[tb]
    \centering
    \includegraphics[width=\linewidth]{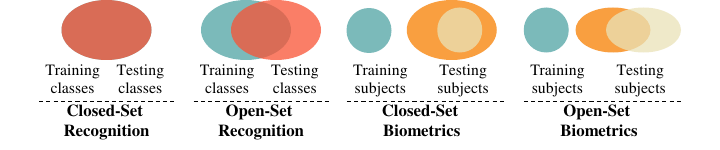}
    \vspace{-0.1in}
    \caption{Comparison between open-set recognition and biometrics. Closed-set recognition has the same set of classes for training and testing. Open-set recognition involves testing on unseen classes during training. 
    On the other hand, in biometrics, the sets of training and testing subjects are always disjoint. In closed-set biometrics, all probe subjects are in the gallery, and in open-set biometrics, some probe subjects are not enrolled in the gallery.
    [Legends: \textcolor[HTML]{F99F40}{\rule{7.5pt}{7.5pt}} gallery subjects, \textcolor[HTML]{EFE9C9}{\rule{7.5pt}{7.5pt}} probe subjects]}
    \label{fig:problem_compare}
\end{figure}

\subsubsection{Open-Set Recognition.}
While open-set biometrics is relatively underexplored, open-set recognition~\cite{kong2021opengan, cen2023the, zhai2023soar, yang2022openood, vaze2022openset, wang2022openauc, bendale2016towards, huang2022class, kuchibhotla2022unseen, zhang2020hybrid, zhou2021learning, yoshihashi2019classification, koch2023lord, chen2021adversarial, ren2024chatgpt} has attracted attention in the literature. Open-set recognition distinguishes between known and unknown classes in image classification. 
In both open-set image recognition and open-set biometrics, the primary concern is how the models deal with instances that belong to unknown classes or non-mated probes. 
However, as illustrated in \cref{fig:problem_compare}, in open-set recognition known classes come from the training set and the unknown classes refer to classes not used for training~\cite{geng2020recent}.
On the other hand, even in closed-set biometrics, all test subjects are typically not used during training and mated probes belong to test subjects enrolled in the gallery~\cite{jain2021biometrics, minaee2023biometrics}. Open-set biometrics differs from closed-set biometrics in that there are non-mated subjects are neither in the training set nor in the gallery~\cite{open_set_fingerprint, open-set_lfw, frvt}.
In evaluation, open-set recognition models typically use closed-set recognition accuracy and AUROC for detecting classes not seen during training~\cite{geng2020recent} and open-set biometrics uses FNIR@FPIR~\cite{frvt}.

%% file: paper/3_method.tex
\section{Method}

\subsection{Problem Formulation}

\begin{figure}[t!]
    \centering
    \includegraphics[width=0.85\linewidth]{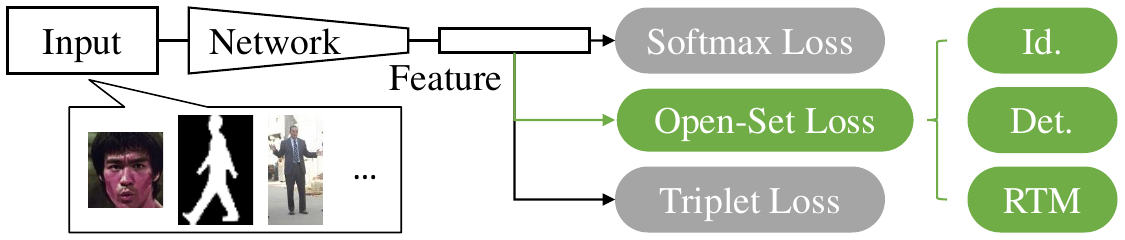}
    \caption{Overview of our approach. We introduce an additional modality-agnostic loss function specifically designed for open-set biometrics, which simultaneously optimizes identification (Id.) and detection (Det.) and Relatie Threshold Minimization (RTM) for open-set thresholds.}
    \label{fig:flow_overview}
\end{figure}

Let \(\mathcal{S}_{\text{test}}\) be the test set, $\mathcal{G} \subset \mathcal{S}_{\text{test}}$ represent the gallery set containing mated subjects, and $\mathcal{P} \subset \mathcal{S}_{\text{test}}$ denote the probe set consisting of test samples. 
For each probe sample $p_i \in \mathcal{P}$, there are two possible outcomes: (1) If $p_i$ is mated, \ie, its corresponding subject $g_i$ is in $\mathcal{G}$, the biometric model must identify the correct subject. (2) If $p_i$ is non-mated, the model must reject it. 
The open-set scenario poses a more challenging problem than the traditional closed-set one, as the model needs to recognize mated subjects while rejecting non-mated ones.

\subsection{Revisiting Loss Functions for Biometrics}

In this section, we compute gradients of two common loss functions in biometrics: softmax and triplet loss. 
Many loss functions are derived from them and share similar characteristics. 
For instance, center loss~\cite{center_loss} can be analyzed similarly to softmax loss, and the logic of triplet loss can also be applied to circle loss~\cite{circle_loss}.

\subsubsection{Softmax loss.}

\begin{figure}[t]
    \centering
    \includegraphics[width=\linewidth]{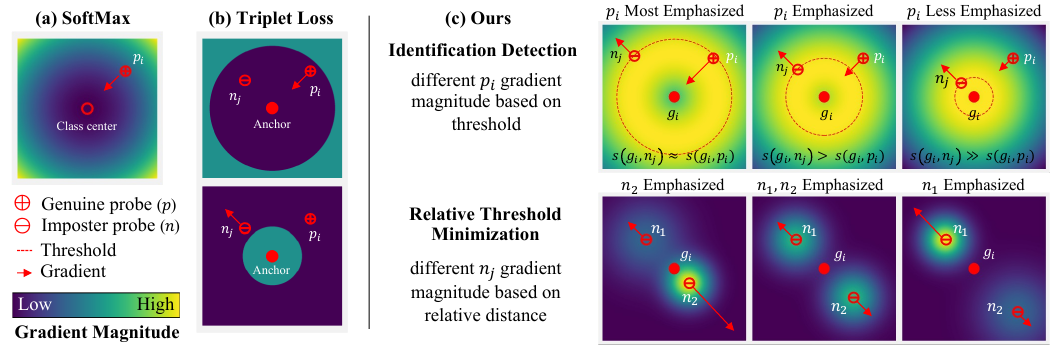}
    \caption{Visualization of the gradient of different loss functions. The colors represent gradient magnitude and the arrows represent gradient directions. a) SoftMax loss pushes the sample toward the class center and away from the other classes, with gradual shifts in gradient magnitude. b) On the other hand, triplet loss has a hard cut gradient based on triplet samples. c) For $S_{\tau}^{det}$, as shown in the top row, the thresholds are determined by non-mated sample, \(n_j\). \(p_i\)'s gradient has the greatest magnitude when it has a similar distance from the gallery \(g_i\) to \(n_j\).
    For Relative Threshold Minimization, as shown in the bottom row, as \(n_2\) moves away from the gallery, its gradient decreases. While \(n_1\) remains at the same location, its gradient increases because it becomes closer to \(g_i\) than \(n_2\).
    The gradients {\it \wrt} genuine scores adapt to non-mated scores and the gradients {\it \wrt} non-mated scores are adaptive to other non-mated scores.}
    \label{fig:loss_grads}
\end{figure}

The softmax loss, often employed in closed-set classification, operates by computing the cross-entropy between the normalized logit of a sample and its corresponding ground truth class.
Mathematically, for a sample \(i\) in a batch, the softmax loss \(L_{\text{softmax}}(i)\) is:
\begin{align}
    L_{\text{softmax}}(i) = -\log\left(\frac{e^{z_{i, s}}}{\sum_{t} e^{z_{i, t}}}\right).
\end{align}
Here, \(z_{i, s}\) denotes the logit for sample \(i\) belonging to subject \(s\). % and the sum in the denominator goes over all the training subjects. 
This loss aims to minimize the negative log-likelihood of the true subject \(s\) for sample \(i\). 
For one sample, the gradient of the softmax loss {\it w.r.t.}~the logit \(z_{i, s}\) is given by:
\begin{align}
    \frac{\partial L_{\text{softmax}}(i)}{\partial z_{i, s}} = \frac{e^{z_{i, s}}}{\sum_{t} e^{z_{i, t}}} - \delta_{i,s},
\end{align}
where \(\frac{e^{z_{i, s}}}{\sum_{j=1}^{N} e^{z_{j, s}}}\) represents the probability that sample \(i\) belongs to subject \(s\) and \(\delta_{i,c}\) is the Kronecker delta, which is $1$ when \(i = c\) (the true subject) and $0$ otherwise.
We visualize gradients of the softmax loss in \cref{fig:loss_grads}(a). 

However, it overlooks the relative magnitude of each imposter score within the batch. 
An imposter score may be deemed small by, \eg, AdaFace~\cite{adaface} standards but still rank high among maximum-per-probe non-mated scores, raising the open-set threshold. 
Further, the softmax loss assumes that each sample can be assigned to one of the training subjects, which is not the case in open-set settings. 
Consequently, the softmax loss is not inherently designed to handle open-set biometrics.

\subsubsection{Triplet loss.}
The triplet loss~\cite{facenet} is defined as:
\begin{align}
    L_{\text{tri}}(a, p, n) = \max(0, d(a, p) - d(a, n) + m).
\end{align}
Note that $d(a, p)$ is the distance between the anchor $a$ and positive sample $p$, $d(a, n)$ is between $a$ and the negative sample $n$, and $m$ is a constant margin.
The partial derivatives of $L(a, p, n)$ {\it \wrt} $d(a, p)$ and $d(a, n)$ are piecewise functions:
\begin{align}
\frac{\partial L_{\text{tri}}(a, p, n)}{\partial d(a, p)} & = 
\begin{cases}
1, & \text{if } d(a, p) + m > d(a, n) \\
0, & \text{if } d(a, p) + m \leq d(a, n) \\
\end{cases};\\
\frac{\partial L_{\text{tri}}(a, p, n)}{\partial d(a, n)} & = 
\begin{cases}
-1, & \text{if } d(a, p) + m > d(a, n) \\
0, & \text{if } d(a, p) + m \leq d(a, n) \\
\end{cases}.
\end{align}

The magnitude of the gradient is constant regardless of the distance between two samples. This inability to differentiate samples in the gradient implies that all negative samples in the triplet are contributing equally, which is not reflective of the open-set scenario. One remedy often used with triplet loss is hard negative mining. However, it comes with a critical drawback of being susceptible to label noise~\cite{fat}. Noisy and mislabeled samples are often considered hard negatives, which can make the loss function sensitive to noisy data when focusing on the hardest samples. Thus, we believe that a loss function that considers all negative samples while being able to adjust the gradient magnitude based on the anchor-negative distances would be effective in open-set biometrics. 

Hence, we propose a novel loss function, shown in \cref{fig:flow_overview}. 
It optimizes identification and detection at various thresholds and minimizes non-mated scores to reduce open-set thresholds. 
It can be applied to any biometric modality. 
Unlike existing loss functions, it is consistent with FNIR@FPIR in that it treats genuine and imposter scores asymmetrically and tackles large non-mated scores. 

\subsection{Our Approach}

To design loss functions tailored for open-set biometrics, we create a training scenario mirroring the testing environment. During training, we partition each batch into gallery and probe sets to simulate open-set evaluation conditions. In this process, we randomly designate a certain percentage, denoted as \(p\%\), of the subjects as mated subjects, while the rest are categorized as non-mated. For the mated subjects, we further randomly allocate their exemplars to the gallery and probe sets. Conversely, all exemplars of the non-mated subjects are placed in the probe set. This approach yields a more realistic and representative training scenario, preparing the biometric features to excel under open-set conditions. To avoid any potential notation ambiguity, we refer to the mated probe set created within each training batch as \(\mathcal{P}_\text{K}'\), the non-mated probe set as \(\mathcal{P}_\text{U}'\), and the gallery sets within each training batch as \(\mathcal{G}'\).

To ensure robustness, when the metric space inhabited by the biometric features is defined under the Euclidean distance, we calculate the similarity scores between two features as:
\begin{align}
\operatorname{s}(p_i, g_j) = \frac{1}{1 + \operatorname{d}(p_i, g_j)},
\end{align}
where $\operatorname{d}(\cdot, \cdot)$ represents the Euclidean distance. Alternatively, if the metric space is defined under the cosine distance, we simply adopt the cosine similarity.

To ensure that the model can generalize to open-set scenarios during testing, we propose loss functions that penalize the model for three types of errors during training: (1) failing to detect a mated probe with a threshold $\tau$, (2) failing to identify a mated probe within the top rank-$r$ positions and (3) setting the threshold $\tau$ too high, leading to poor differentiation between mated and non-mated sample distributions.

\subsubsection{(1) Detection.}
Detection is assessing whether the computed similarity score exceeds a predefined threshold $\tau$.
That is, for a given probe \(p_i \in \mathcal{P}_\text{K}'\) and its corresponding gallery subject \(g_i \in \mathcal{G}'\), we calculate a detection score as:
\begin{align}
    S^{det}_\tau(p_i, g_i) = \sigma_\alpha\left(\operatorname{s}(p_i, g_i) - \tau\right),
\end{align}
where \(\sigma_\alpha(x) = 1 / (1 + \exp(-\alpha x))\) is the Sigmoid function with temperature hyperparameter \(\alpha\).
The \(S^{det}_\tau(p_i, g_i)\) score quantifies the model's ability to detect if \(p_i\) and \(g_i\) belong to the same subject, given the threshold \(\tau\). A high similarity score, well above \(\tau\), indicates successful detection and \(S^{det}_\tau(p_i, g_i)\) approaches 1. In contrast, if the model fails to detect that \(p_i\) and \(g_i\) belong to the same subject, \(s(p_i, g_i)\) remains much smaller than \(\tau\), causing \(S^{det}_\tau(p_i, g_i)\) to approach 0. Since the Sigmoid function has the highest gradient near $0$, the term $\operatorname{s}(p_i, g_i) - \tau$ makes the loss function focus more on the samples near the threshold $\tau$. 

During testing, the open-set detection threshold is determined by the non-mated score distribution, which we do not know during training. Thus, we sample \(\tau\) from the non-mated scores in a batch. Specifically, we let
\begin{align}
    S^{\mathit{det}} = \frac{1}{\left\lvert \mathcal{T} \right\rvert}\sum_{\tau \in \mathcal{T}} S^{det}_\tau(p_i, g_i),
    \label{eq:det}
\end{align}
where \(\mathcal{T} = \left\{s(n_j, g_i) \vert n_j \in \mathcal{P}_\text{U}' \right\}\) is the set of all non-mated scores.
\cref{fig:loss_grads}(c) illustrates the gradient of $\operatorname{Det}$ {\it \wrt}~mated probes under different thresholds induced by various non-mated probes. The gradient is adaptable to different thresholds, increasing as $\tau$ decreases, even when $s(g, p)$ is constant. This adaptability is desirable in open-set settings, as it promotes mated probes to be more similar to their ground-truth gallery than non-mated probes.
Note that the thresholds in \cref{eq:det} are only used during training. The detection thresholds during testing are automatically determined by the FNIR@FPIR metric as discussed in \cref{sec:intro}.

\subsubsection{(2) Identification.}
It is not sufficient for a biometric model to distinguish mated subjects from non-mated ones based on their scores. % T
Equally important is the model's ability to accurately identify the correct subject among the gallery subjects. Therefore, we introduce an identification score $S^{id}(p_i, g_i)$  as:
\begin{align}
    S^{id}(p_i, g_i) = \sigma_\beta(1 - \operatorname{softrank}(p_i, g_i)),
\end{align}
where \(\sigma_\beta\) is a Sigmoid function with temperature \(\beta\) and \(\operatorname{softrank}\) is a differentiable function with a Sigmoid function \(\sigma_\gamma\) defined as:
\begin{align}
    \operatorname{softrank}(p_i, g_i) = \sum_{g_j \in \mathcal{G}'} \sigma_\gamma(\operatorname{s}\left(p_i, g_j\right) - \operatorname{s}\left(p_i, g_i\right)),
\end{align}

The \(\operatorname{softrank}\) function accumulates values close to 1 for gallery subjects \(g_j\) more similar to the probe \(p_i\) than the correct match \(g_i\) and close to 0 otherwise. It essentially reflects the rank of the correct match \(g_i\) among all gallery subjects for probe \(p_i\). The \(S^{id}(p_i, g_i)\) is an indicator, approaching 1 if \(p_i\) has a \(\operatorname{softrank}\) value less than \(1\), indicating successful identification.

To optimize open-set detection and identification simultaneously, the overall identification-detection loss \(\mathcal{L}_{r, \tau}\) is formulated as the product of two components,
\begin{align}
    \mathcal{L}^{\operatorname{IDL}} = -\frac{1}{\left\lvert \mathcal{P}_\text{K}' \right \rvert} \sum_{p_i \in \mathcal{P}_\text{K}'} S^{det}(p_i, g_i) \cdot S^{id}(p_i, g_i).
    \label{eq:idl}
\end{align}
In other words, $\mathcal{L}^{\operatorname{IDL}}$ is only close to $-1$ if both \(S^{id}(p_i, g_i)\) and \(S^{det}(p_i, g_i)\) approach $1$. Conversely, should either or both not approximate $1$, $\mathcal{L}^{\operatorname{IDL}}$ converges to $0$.

\subsubsection{(3) Relative Threshold Minimization}

$\mathcal{L}^{\operatorname{IDL}}$ optimizes detection and identification under fixed thresholds. We also introduce a strategy to minimize the maximum scores for non-mated probes, discouraging the model from assigning high scores to non-mated probes. This helps mitigate the risk of false positive detections in open-set scenarios, enhancing overall robustness.

To achieve this, we minimize the weighted average of the similarities between the probe and each gallery subject, where the weights are obtained by applying the softmax function to each score. 
Mathematically, given a probe sample $p_i$ and a set of gallery subjects $\mathcal{G}'= \{g_1, g_2, ..., g_n\}$ with corresponding similarity scores $s_1, s_2, ..., s_n$, we minimize the weighted average as follows:
\begin{align}
    \mathcal{L}^{\text{RTM}} = \frac{1}{\sum_{j=1}^{n} e^{s_j}} \sum_{j=1}^{n} e^{s_j} \cdot s_j,
    \label{eq:rtm}
\end{align}
where $e^{s_j}$ is the softmax value for each similarity score. 

\cref{fig:loss_grads} (c) shows gradients of $\mathcal{L}_{\text{RTM}}$ {\it \wrt}~non-mated probes. 
This gradient adapts to the relative position of non-mated probes, assigning higher gradients to samples closer to the gallery. 
Unlike softmax-based losses and the triplet loss, RTM consistently emphasizes imposter scores that rank higher compared to others within the batch to learn better feature representations for open-set biometrics.
This property is desirable as it effectively minimizes the maximum score of each non-mated probe.

The \verb|max| function is an alternative to the weighted average. 
Yet, simply minimizing the maximum score in training would not minimize other scores even if they are close to the maximum. 
Hence, the weighted average would make it easier for the trained model to generalize during testing.

\subsection{Overall Loss}

The overall objective is to minimize the combination of $ \mathcal{L}^{\operatorname{IDL}}$ and $\mathcal{L}^{\text{RTM}}$, %is denoted by the hyperparameter \(\lambda\),
\begin{align}
    L = \mathcal{L}^{\text{IDL}} + \lambda \cdot \mathcal{L}^{\text{RTM}}.
\end{align}
where \(\lambda\) is the loss weight.
Compared to existing loss functions, our loss function treats genuine and imposter scores differently. It aims to reduce open-set thresholds by pushing non-mated probes away from gallery while pulling genuine scores toward their corresponding galleries, similar to open-set evaluation. Moreover, the relative magnitude of non-mated scores is utilized by RTM to assign larger gradients to harder non-mated probes.

%% file: paper/4_experiments.tex
\section{Experiments}

\subsection{Experimental Setup}

\subsubsection{Evaluation Protocols.}
In face recognition experiments, we use the official closed- and open-set evaluation protocols of IJB-S~\cite{ijb-s}. However, gait recognition and person reID datasets typically lack official open-set evaluation protocols, where some probe subjects are excluded from the gallery. Na\"{i}vely removing gallery subjects often results in highly unstable outcomes. This instability arises because threshold calculation heavily relies on a small fraction (1\%) of non-mated probes due to the limited number of probes in these datasets. For instance, in a dataset with $500$ non-mated probes, the threshold is computed from the highest scores of only $5$ probes. 

To ensure robust open-set evaluation, we create $N$ ($=\!50$) random sets of non-mated probes and compute FNIR@1\%FPIR for each set. In each set, \(q\%=21.5\%\) subjects are designated as non-mated.
We report the median value and the standard deviation from $N$ trials. 
The choice of median over mean is deliberate to avoid undue influence from extreme values and provide a more stable metric.

\begin{wraptable}[14]{r}{0.55\textwidth}
    \vspace{-22pt}
    \caption{Summary of datasets used in our experiments. 
    [Keys: Mod.=Modality; Img.=Open-set image recognition; Subj.=the number of subjects \iffalse for biometric datasets and the number of classes for miniImageNet \fi; Media=the number of images for face/image recognition datasets and the number of sequences/tracklets for gait and video-based person reID datasets]
    }
    \label{tab:datasets}
    \centering\footnotesize
    \resizebox{\linewidth}{!}{
    \begin{NiceTabular}{lccrrrr}
        \toprule
        \Block{2-1}{Dataset} & \Block{2-1}{Year} & \Block{2-1}{Mod.} & \Block{1-2}{Train} & & \Block{1-2}{Test} & \\ \cmidrule(lr){4-5} \cmidrule(lr){6-7}
        & & & Subj. & Media & Subj. & Media \\ \midrule
        IJB-S~\cite{ijb-s} & 2018 & Face & - & - & 202 & $>10$M \\ 
        Gait3D~\cite{gait3d} & 2022 & Gait & \num{3000} & \num{18940} & \num{1000} & \num{6369} \\ 
        GREW~\cite{grew} & 2021 & Gait & \num{20000} & \num{102887} & \num{6000} & \num{24000} \\ 
        MEVID~\cite{mevid} & 2023 & ReID & \num{104} & \num{6338} & \num{55} & \num{1754} \\
        CCVID~\cite{cal} & 2022 & ReID & \num{75} & \num{948} & \num{151} & \num{1908} \\
        \bottomrule
    \end{NiceTabular}
    }
\end{wraptable}

\subsubsection{Evaluation Metrics.}
We evaluate in both open- and closed-set settings. 
In closed-set settings, we use the Cumulative Matching Curve (CMC) at rank-1; for open-set, we use FNIR@FPIR at \(1\%\) and \(0.1\%\) FPIR and rank $R=20$.

\subsubsection{Datasets.}
For face recognition, we train on WebFace4M~\cite{webface260m} and test on IJB-S~\cite{ijb-s}. Our experiments in gait recognition involve the Gait3D~\cite{gait3d} and GREW~\cite{grew} datasets, while for person reID, we employ the MEVID~\cite{mevid} and CCVID~\cite{cal} datasets. 
Further details are provided in \cref{tab:datasets}.

\label{sec:exp_setup}
One technical caveat is that although FNIR@1\%FPIR implicitly assumes that each subject only appears once in the gallery, a typical gait recognition or person re-id dataset contains more than one gallery template per subject. To address this issue, we take the mean feature over all gallery templates for each subject. This benefits the models compared to randomly sampling one gallery template (details are in the Supp.).

\subsubsection{Implementation Details.}

We implement our proposed approach in PyTorch~\cite{pytorch}. For fair comparisons, we follow the experiment setup of our baselines. For details on the hyperparameters, please refer to the Supp.

\subsection{Ablation Studies}

We conduct our ablation studies on the Gait3D~\cite{gait3d} dataset using GaitBase~\cite{opengait} as the baseline. In the ablation studies, we hold out \(20\%\) of subjects in the training set as the validation set. 

\begin{wraptable}[14]{r}{0.55\textwidth}
    \vspace{-25pt}
    \caption{Ablation study results. [Keys: \ding{51}=Kept; \ding{55}=Removed;    R@1=rank-1 accuracy]}
    \label{tab:ablation_study}
    \centering\scriptsize
    \resizebox{\linewidth}{!}{
    \begin{NiceTabular}{cccccccc}
    \CodeBefore
        \rectanglecolor{red!15}{3-1}{4-3}
        \rectanglecolor{red!15}{5-4}{6-4}
        \rectanglecolor{red!15}{7-5}{8-5}
        \rectanglecolor{red!15}{9-6}{10-6}
        \rectanglecolor{green!15}{11-1}{11-8}
    \Body
        \toprule
        \(\operatorname{Id}\) & \(\operatorname{Det}\) & RTM & \(\alpha, \gamma\) & \(\beta\) & \(\lambda\) & R@1 & FNIR@1\%FPIR \\
        \midrule
        \ding{55} & \ding{55} & \ding{55} & --- & --- & --- & \(64.6\) & \(86.43 \smallpm 5.44\) \\ \midrule
        \ding{55} & \ding{55} & \ding{51} & \Block{2-1}{$6.0$} & \Block{2-1}{$0.2$} & \Block{2-1}{$4.0$} & \(65.1\) & \(85.81 \smallpm 5.78\) \\
        \ding{55} & \ding{51} & \ding{51} & & & & \(64.4\) & \(85.16 \smallpm 5.13\) \\ \midrule
        \Block{2-1}{\ding{51}} & \Block{2-1}{\ding{51}} & \Block{2-1}{\ding{51}} & $3.0$ & \Block{2-1}{$0.2$} & \Block{2-1}{\(4.0\)} & \(63.9\) & \(85.22 \smallpm 4.73\) \\
         & & & $9.0$ & & & \(64.2\) & \(84.08 \smallpm 6.28\) \\ \midrule
        \Block{2-1}{\ding{51}} & \Block{2-1}{\ding{51}} & \Block{2-1}{\ding{51}} & \Block{2-1}{$6.0$} & $0.1$ & \Block{2-1}{$4.0$} & \(63.1\) & \(85.10 \smallpm 5.67\) \\
        & & & & $0.4$ & & \(63.8\) & \(85.92 \smallpm 5.45\) \\ \midrule
        \Block{2-1}{\ding{51}} & \Block{2-1}{\ding{51}} & \Block{2-1}{\ding{51}} & \Block{2-1}{$6.0$} & \Block{2-1}{$0.2$} & \(2.0\) & \(64.8\) & \(83.12 \smallpm 5.94\) \\ 
         & & & & & \(8.0\) & \(61.6\) & \(85.67 \smallpm 5.15\) \\ \midrule
        \ding{51} & \ding{51} & \ding{51} & $6.0$ & $0.2$ & \(4.0\) & \(64.8\) & \(82.99 \smallpm 4.50\) \\
        \bottomrule
    \end{NiceTabular}
    }
\end{wraptable}

\subsubsection{Effect of \(\operatorname{Det}\), \(\operatorname{Id}\) and RTM.}
We ablate the effect of \(\operatorname{Det}\), \(\operatorname{Id}\) and RTM in \cref{tab:ablation_study}. 
RTM consistently enhances both closed- and open-set performance. \(\operatorname{Id}\) further improves open-set performance but has a minor impact on rank-1 accuracy. Meanwhile, \(\operatorname{Det}\) significantly enhances both closed- and open-set performance. In summary, each component contributes to its effectiveness.

\subsubsection{Effect of Hyperparameters.}
We ablate the effect of hyperparameters \(\alpha, \beta, \gamma,\) and \(\lambda\) in \cref{tab:ablation_study}. Across different hyperparameter settings, our approach can consistently improve open-set performance compared to the baseline, GaitBase~\cite{opengait}. Also, while the closed-set rank-1 accuracy is more sensitive to the hyperparameters, our approach can achieve slightly higher rank-1 accuracy with the best set of tested hyperparameters.

\subsection{Comparison with State of the Art}

Across biometric modalities, existing methods exhibit competitive closed-set recognition performance, as shown in \cref{tab:ijb-s_exp} for face, \cref{tab:gait3d_exp} for gait, and \cref{tab:combined_exp} for person reID. 
In particular, in gait recognition, GaitBase~\cite{opengait} and SwinGait~\cite{swingait} showcase strong closed-set performance with a notable rank-1 accuracies of greater than \(60\%\) and \(70\%\), respectively, for in-the-wild gait datasets Gait3D~\cite{gait3d} and GREW~\cite{grew}.
However, in open-set scenarios, they face challenges, as evidenced by GaitBase's high FNIR of \(86.43\%\) at \(1\%\) FPIR in Gait3D. %~\cite{gait3d}.

\begin{table}[tb]  % [10]{R}{0.6\textwidth}
    \caption{Closed- and open-set face recognition on two official protocols of IJB-S~\cite{ijb-s}. %\TODO{update the results}
    [Keys: R@1=rank-1 accuracy; 0.1\%=FNIR@0.1\%FPIR; 1\%: FNIR@1\%FPIR]}
    \label{tab:ijb-s_exp}
    \centering\scriptsize
    \resizebox{0.6\linewidth}{!}{
    \begin{NiceTabular}{lcccccc}
        \toprule
        \Block{2-1}{Model} & \Block{1-3}{Surveillance-Single} & & &  \Block{1-3}{Surveillance-Booking} \\ \cmidrule(lr){2-4} \cmidrule(lr){5-7}
        & R@1 $\uparrow$ & 0.1\% $\downarrow$ & 1\% $\downarrow$ & R@1 $\uparrow$ & 0.1\% $\downarrow$ & 1\% $\downarrow$ \\ \midrule
        AdaFace~\cite{adaface} & $68.44$ & $45.18$ & $38.37$ & $68.84$ & $45.98$ & $37.80$ \\
        AdaFace+Ours & \(69.92\) & \(44.30\) & \(36.73\) & \(70.00\) & \(43.48\) & \(36.17\) \\
        \bottomrule
    \end{NiceTabular}
    }
\end{table}

It is evident that the inclusion of our loss functions consistently yields improvements in both closed- and open-set performance across three modalities. 
In face recognition, we boost the rank-1 accuracy of AdaFace~\cite{adaface} to \(69.92\%\) (\(+1.48\%\)) in Surveillance-to-Single. 
In the challenging open-set scenario, our losses further demonstrate their effectiveness by reducing FNIR@0.1\%FPIR from \(45.98\%\) to \(43.48\%\) in Surveillance-Booking. 

\begin{wraptable}[16]{R}{0.6\textwidth}
    \vspace{-22pt}
    \caption{Closed- and open-set gait recognition on Gait3D~\cite{gait3d} and GREW~\cite{grew}. SMPLGait~\cite{gait3d} takes SMPL parameters as additional input. SwinGait-3D~\cite{swingait} uses our own implementation. [Keys: Rank@1=rank-1 accuracy]}
    \label{tab:gait3d_exp}
    \centering\scriptsize
    \resizebox{\linewidth}{!}{
    \begin{NiceTabular}{lcccc}
        \toprule
        \Block{2-1}{Model} & \Block{1-2}{Rank@1 $\uparrow$} & & \Block{1-2}{FNIR@1\%FPIR $\downarrow$} \\ \cmidrule(lr){2-3} \cmidrule(lr){4-5}
        & Gait3D & GREW & Gait3D & GREW \\ \midrule
        GaitSet~\cite{gaitset} & $36.7$ & $48.4$ & $96.05 \smallpm 2.30$ & $81.17 \smallpm 1.45$ \\
        GaitPart~\cite{gaitpart} & $28.2$ & $47.6$ & $96.88 \smallpm 1.80$ & $88.43 \smallpm 1.26$ \\
        GaitGL~\cite{gaitgl} & $29.7$ & $41.5$ & $94.97 \smallpm 1.24$ & $80.55 \smallpm 1.11$ \\
        SMPLGait~\cite{gait3d} & $46.3$ & --- & $93.18 \smallpm 2.58$ & ---\\ \midrule
        GaitBase~\cite{opengait} & $64.6$ & $60.1$ & $86.43 \smallpm 5.44$ & $77.67 \smallpm 1.60$ \\
        \quad +Ours & $64.8$ & \(60.1\) & $82.99 \smallpm 4.50$ & \(76.55 \smallpm 1.54\) \\ \midrule
        SwinGait-3D~\cite{swingait} & $72.4$ & $79.3$ & $83.50 \smallpm 6.20$ & \(58.32 \smallpm 1.79\) \\
        \quad +Ours & $75.3$ & \(80.4\) & $79.30 \smallpm 6.74$ & \(57.29\smallpm 1.83\) \\
        \bottomrule
    \end{NiceTabular}
    }
\end{wraptable}

In gait recognition, when applied to Swin-Gait~\cite{swingait} backbone, our losses improve the rank-1 accuracy to \(75.3\%\) (+\(2.9\%\)) on Gait3D, and FNIR@1\%FPIR decreases to \(79.30\%\) (-\(4.20\%\)) on Gait3D. 
In person reID,  integrating our loss with CAL~\cite{cal} reaches a rank-1 accuracy of \(84.8\%\) (\(+2.2\%\)) and a FNIR of \(30.81\%\) (\(-3.08\%\)) on MEVID~\cite{mevid}. 
We conduct the independent two-sample t-test to assess the statistical significance of performance improvements in open-set scenarios. For instance, our approach outperforms CAL on the CCVID~\cite{cal} dataset, with a p-value of \(0.052\%\).

Collectively, these findings emphasize the consistent effectiveness of our proposed loss functions in enhancing both closed- and open-set performance across a spectrum of biometric tasks, ultimately improving the models' ability to handle open-set challenges, while maintaining strong closed-set performance.

\begin{table}[t!]
    \caption{Closed- and open-set person reID on MEVID~\cite{mevid} and CCVID~\cite{cal}.}
    \label{tab:combined_exp}
    \centering\scriptsize
    \begin{NiceTabular}{lccc}
        \toprule
        Model & Dataset & Rank-1 Accuracy $\uparrow$ & FNIR@1\%FPIR $\downarrow$ \\ \midrule
        AGRL~\cite{agrl} & MEVID~\cite{mevid} & $48.4$ & $66.67 \smallpm 9.57$ \\
        TCLNet~\cite{tclnet} & MEVID~\cite{mevid} & $48.1$ & $70.34 \smallpm 11.68$ \\
        PSTA~\cite{psta} & MEVID~\cite{mevid} & $48.2$ & $77.11 \smallpm 11.72$ \\ \midrule
        CAL~\cite{cal} & MEVID~\cite{mevid} & $52.5$ & $66.95 \smallpm 11.59$ \\
        \quad +Ours & MEVID~\cite{mevid} & $55.1$ & $64.80 \smallpm 10.71$ \\ \midrule
        CAL~\cite{cal} & CCVID~\cite{cal} & $82.6$ & \(33.89 \smallpm 7.65\) \\
        \quad +Ours & CCVID~\cite{cal} & $84.8$ & \(30.81 \smallpm 8.02\) \\
        \bottomrule
    \end{NiceTabular}
\end{table}

\subsubsection{Efficiency.}

\begin{table}[tb]  % [11]{r}{0.55\textwidth}
    \caption{FLOPs in training different baseline models with and without our approach. 
    We report the mean FLOPs across a minibatch. Our approach incurs no additional FLOP during testing.}
    \label{tab:flops_results}
    \centering\scriptsize
    \resizebox{0.55\linewidth}{!}{
    \begin{NiceTabular}{ccrrr}
        \toprule
        Modality & Model & FLOPS & +Triplet & +Ours \\ \midrule
        Face & AdaFace~\cite{adaface} & $24.19$G & --- & $503$K \\
        Gait & GaitBase~\cite{opengait} & $35.43$G & $524.3$K & $245.8$K \\
        Gait & SwinGait~\cite{swingait} & $120$G & $491.5$K & $230.4$K \\
        Body & CAL~\cite{cal} & $32.75$G & $4.09$K & $15.38$K \\
        \bottomrule
    \end{NiceTabular}
    }
\end{table}

We present the floating point operations (FLOPs) of training baselines with and without our loss functions in \cref{tab:flops_results}. 
It is expected that incorporating our loss into baselines will increase the computational load in training, but the increase is very minor.
For example, when SwinGait~\cite{swingait} is used, our loss functions only incur \(46.9\%\) of the FLOPs of the triplet loss, which is negligible compared to the model.
Moreover, our approach does not incur any computation overhead during testing while improving closed- and open-set performance.

\subsection{Visualizations}

\subsubsection{Score Distributions.}

\begin{wrapfigure}[20]{r}{0.6\textwidth}
    \centering
    \includegraphics[width=\linewidth]{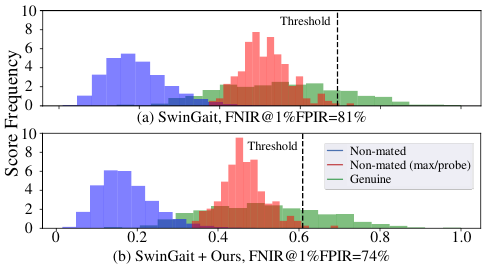}
    \caption{Score distributions on the test set of Gait3D~\cite{gait3d} dataset. 
    Ours (b) effectively reduces the threshold at $1\%$ FPIR while only shifting the genuine score distribution marginally to the left compared to SwinGait~\cite{swingait} (a). 
    Both distributions are shifted and scaled horizontally so that the min/max scores are within $[0,1]$.
    }
    \label{fig:score_dist}
\end{wrapfigure}

\cref{fig:score_dist} visualizes the non-mated, maximum non-mated score for each probe, and genuine scores of SwinGait~\cite{swingait},  with and without our loss. 
Our introduced RTM effectively reduces the distribution of maxi-mum-per-probe scores, shifting it towards lower values. 
After linearly normalizing all scores between $0$ and $1$, the threshold at $1\%$ FPIR is $0.61$ with our loss (\cref{fig:score_dist} (b)) compared to $0.69$ without ours (\cref{fig:score_dist} (a)). 
Moreover,  our loss does not significantly alter the rightmost portion of the genuine score distribution, enhancing detection.
This analysis offers insights into how our loss changes score distributions, and through which improves open-set performance.

\subsubsection{False Negatives Breakdown.}

\begin{figure}[tb]
    \centering
    \begin{subfigure}{0.32\linewidth}
        \includegraphics[width=\linewidth]{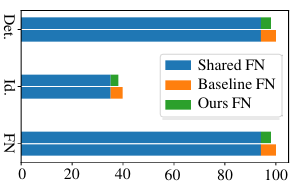}
        \caption{Face}
        \label{subfig:face_fn}
    \end{subfigure}
    \hfill
    \begin{subfigure}{0.32\linewidth}
        \includegraphics[width=\linewidth]{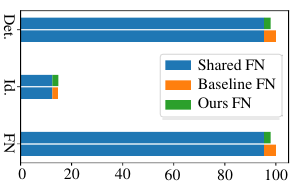}
        \caption{Gait}
        \label{subfig:gait_fn}
    \end{subfigure}
    \hfill
    \begin{subfigure}{0.32\linewidth}
        \includegraphics[width=\linewidth]{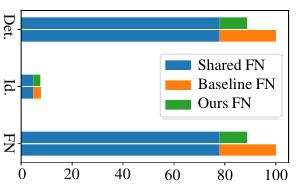}
        \caption{Person ReID}
        \label{subfig:person_reid_fn}
    \end{subfigure}
    \caption{Open-set FNs due to detection, identification, and either. The average across the $50$ open-set evaluation protocols is plotted. Blue indicates FNs that are shared by the baseline, SwinGait~\cite{swingait}, and our approach. Orange indicates FNs of the baseline only. Green indicates FNs of our approach.} 
    \label{fig:gait_fn_breakdown}
\end{figure}

As illustrated in \cref{fig:gait_fn_breakdown}, across different biometric and generic tasks, almost all of the false negatives are failed detection, {\it e.g.}~the $1\textsuperscript{st}$ mated probe in \cref{fig:fnir_fpir}. 
A small portion of false negatives have both failed identification and detection, {\it e.g.}~the $3\textsuperscript{rd}$ mated probe in \cref{fig:fnir_fpir}. 
When we apply our loss functions, the number of FNs due to detection is reduced and the number of FNs due to identification is comparable to that of the baseline. 
These results highlight that our approach can improve open-set detection without sacrificing identification in various settings.

%% file: paper/5_conclusions.tex
\section{Conclusions}

Our work addresses the critical gap in open-set biometrics, a practical generalization of the well-studied closed-set biometrics. 
We introduce novel I-DL and RTM losses aiming to improve biometric models' open-set performance. 
Through extensive experiments across diverse modalities, including face, gait, and person reID, we demonstrate the efficacy of our approach. 
We not only advance the state of the art in the challenging open-set domain but also enhance closed-set recognition capabilities. 
We advocate for increased attention to the open-set problem within the biometric community.

\subsubsection{Limitations}
While our paper makes significant strides in addressing open-set challenges in biometrics through loss functions, our exploration does not extensively explore dedicated architectures for this specific context. Future research endeavors may benefit from investigating tailored architectures that complement and enhance the effectiveness of our proposed loss functions.

\subsubsection{Potential Societal Impacts.}
The use of biometric datasets in our work containing personally identifiable images prompts a crucial discussion on privacy and ethical implications. As biometric technologies become more prevalent, it is imperative to ensure that the collection of such datasets is aligned with ethical standards and privacy regulations.

\subsubsection{Acknowledgements.}
This research is based upon work supported in part by the Office of the Director of National Intelligence (ODNI), Intelligence Advanced Research Projects Activity (IARPA), via 2022-21102100004. The views and conclusions contained herein are those of the authors and should not be interpreted as necessarily representing the official policies, either expressed or implied, of ODNI, IARPA, or the U.S. Government. The U.S. Government is authorized to reproduce and distribute reprints for governmental purposes notwithstanding any copyright annotation therein.

%% file: paper/9_supp.tex
% \clearpage
% \setcounter{page}{1}
% \maketitlesupplementary

% \setcounter{section}{5}
% \setcounter{table}{6}
% \setcounter{figure}{6}

\begin{figure}[h]
    \centering 
    \begin{subfigure}[b]{0.32\linewidth}
        \includegraphics[width=\linewidth]{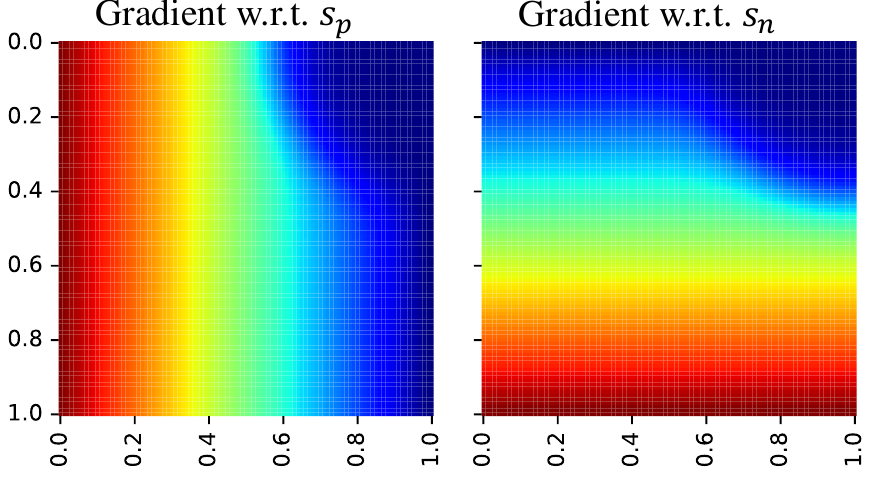}
        \caption{Circle loss.}
        \label{fig:circle_grad}
    \end{subfigure}
    \begin{subfigure}[b]{0.32\linewidth}
        \includegraphics[width=\linewidth]{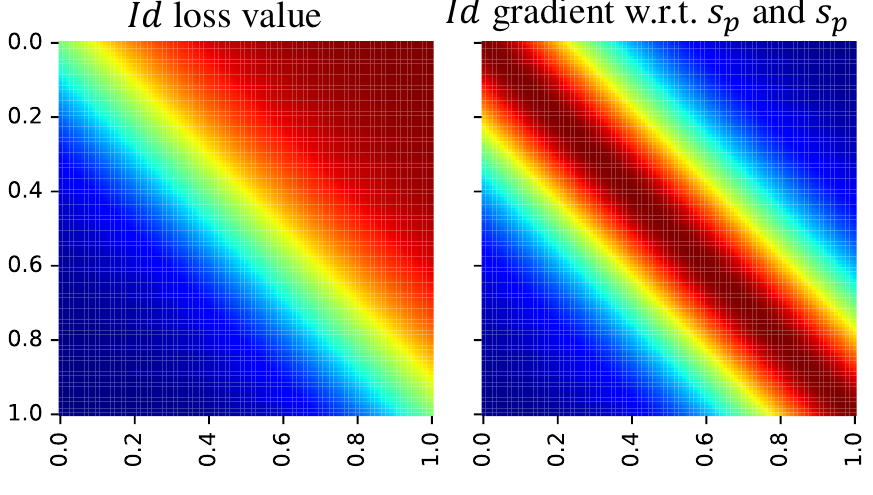}
        \caption{\(\operatorname{Id}\) (Ours).}
        \label{fig:id_val_grad}
    \end{subfigure}
    \caption{\small Comparison between \(\operatorname{Id}\) and Circle loss. The gradient of \(\operatorname{Id}\) \wrt \(s_p\) and \(s_n\) have the same magnitude and opposite directions. [The color scale is the same as Fig.~\textcolor{red}{4}; \(x\)-axes: positive score \(s_p\); \(y\)-axes: negative score \(s_n\)]
    }
    \label{fig:open_set_vs_ciecle}
\end{figure}

\section{Comparison with Other Losses.}

\(\operatorname{Id}\) differs significantly from Circle Loss, MagFace, and InfoNCE, which push all genuine scores to $1$ and imposter scores to $0$. \(\operatorname{Id}\) specifically targets hard-to-classify pairs, enhancing open-set identification by focusing on closely matched genuine and imposter scores. These approaches are compared in \cref{fig:open_set_vs_ciecle}, emphasizing that, unlike existing loss, \(\operatorname{Id}\) is well aligned with open-set challenges.

\section{Additional Experiments}

\subsection{Implementation Details}

For face recognition and person reID, we use the official implementation of AdaFace\footnote{\url{https://github.com/mk-minchul/AdaFace}} and CAL\footnote{\url{https://github.com/guxinqian/Simple-CCReID}}, respectively. 
For gait recognition, we use the official implementation of GaitBase in OpenGait\footnote{\url{https://github.com/ShiqiYu/OpenGait}} and our own implementation of SwinGait~\cite{swingait} based on OpenGait.
\cref{tab:hyperparam} lists the implementation details and hyperparameters in our experiments. Note that to ensure fair comparison in our experiments, we adhere to the original design of our baseline models. Specifically, face and person reID models (AdaFace~\cite{adaface} and CAL~\cite{cal}) use cosine similarity, while gait models (GaitBase~\cite{opengait} and SwinGait~\cite{swingait}) use Euclidean distance.

\begin{table*}
    \caption{Hyperparameters in our experiments. [Keys: MSLR=multi-step learning rate scheduler; SGDR=cosine annealing learning rate scheduler~\cite{loshchilov2016sgdr}; Eps=Epochs; Its=Iterations]}
    \label{tab:hyperparam}
    \centering\small
    \resizebox{\textwidth}{!}{
    \begin{NiceTabular}{llcccccccc}
        % \Body 
        \toprule
        \Block{2-1}{Module} & \Block{2-1}{Description} & \Block{2-1}{Symbol} & AdaFace & \Block{1-2}{GaitBase} & & \Block{1-2}{SwinGait} & & \Block{1-2}{CAL} \\ \cmidrule(lr){5-6}  \cmidrule(lr){7-8} \cmidrule(lr){9-10}
        & & & IJB-S & Gait3D & GREW & Gait3D & GREW & MEVID & CCVID \\ \midrule
        \Block{3-1}{Data} & Input sequence length & --- & \(1\) & \(10\)-\(50\) & \(20\)-\(40\) & \(10\)-\(50\) & \(20\)-\(40\) & \(8\) & \(8\) \\
        & Input imagery height & --- & \(112\) & \(64\) & \(64\) & \(64\) & \(64\) & \(256\) & \(256\) \\
        & Input imagery width & --- & \(112\) & \(44\) & \(44\) & \(44\) & \(44\) & \(128\) & \(128\) \\ \midrule
        \Block{12-1}{Training} & Batch Size & --- & \(1024\) & \(128\) & \(128\) & \(128\) & \(128\) & \(32\) & \(32\) \\
        & Number of subjects per batch & --- & \(512\) & \(32\) & \(32\) & \(32\) & \(32\) & \(8\) & \(8\) \\
        & Number of media per subject & --- & \(2\) & \(4\) & \(4\) & \(4\) & \(4\) & \(4\) & \(4\) \\
        & Optimizer & --- & SGD & SGD & SGD & AdamW & AdamW & Adam & Adam \\
        & Momentum &--- & $0.9$  & \(0.9\) & \(0.9\) & --- & --- & \(0.9\) & \(0.9\) \\
        & Learning rate &--- & $0.1$  \(\) & \(0.1\) & \(0.1\) & \num{3e-4} & \num{3e-4} & \num{3.5e-4} & \num{3.5e-4} \\ 
        & Weight decay &--- & \num{5e-4}  \(\) & \num{5e-4} & \num{5e-4} & \num{2e-2} & \num{2e-2} & \num{5e-4} & \num{5e-4} \\ 
        & Learning rate scheduler& ---  & MSLR & MSLR & MSLR & SGDR & SGDR & MSLR & MSLR \\
        & Learning rate milestones 1 &--- & 12 Eps  & \num{20000} Its & \num{80000} Its  & --- & --- & 40 Eps & 40 Eps \\
        & Learning rate milestones 2 & ---&  20 Eps & \num{40000} Its & \num{120000} Its  & --- & --- & 80 Eps & 80 Eps \\
        & Learning rate milestones 3 & --- & 24 Eps &\num{50000} Its & \num{150000} Its  & --- & --- & 120 Eps & 120 Eps \\
        & Half cosine cycle & --- & --- & --- & --- & \num{60000} & \num{150000} Its & --- \\ \midrule
        \Block{5-1}{Ours} & Sigmoid temperature & \(\alpha\) & \(6.0\) & \(6.0\) & \(9.0\) & \(6.0\) & \(9.0\) & \(6.0\) & \(6.0\) \\
        & Sigmoid temperature & \(\beta\) & \(0.2\) & \(0.2\) & \(0.2\) & \(0.2\) & \(0.2\) & \(0.2\) & \(0.2\) \\
        & Sigmoid temperature & \(\gamma\) & \(6.0\) & \(6.0\) & \(9.0\) & \(6.0\) & \(9.0\) & \(6.0\) & \(6.0\) \\
        & Ratio of non-mated subjects & \(p\) & \(25\%\) & \(25\%\) & \(25\%\) & \(25\%\) & \(25\%\) & \(25\%\) & \(25\%\) \\
        & RTM loss weight & \(\lambda\) & \(4.0\) & \(4.0\) & \(4.0\) & \(4.0\) & \(4.0\) & \(4.0\) & \(4.0\) \\
        \bottomrule
    \end{NiceTabular}
    }
\end{table*}

\subsection{Comparison with EVM}

\begin{table}[tb]
    \caption{Closed- and open-set gait recognition  on Gait3D~\cite{gait3d}.} % [Keys: Rank@1=rank-1 accuracy]}
    \label{tab:evm_comparison}
    \centering\small
    \begin{NiceTabular}{lcc}
        \toprule
        Model & Rank-1 Accuracy $\uparrow$ & FNIR@1\%FPIR $\downarrow$ \\ \midrule
        SwinGait~\cite{swingait} & \(72.4\) & $83.50 \smallpm 6.20$ \\
        \quad +Ours & \(75.3\) & $79.30 \smallpm 6.74$ \\
        \quad +EVM~\cite{open-set_lfw} & \(73.5\) & \(81.08 \smallpm 8.00\) \\
        \bottomrule
    \end{NiceTabular}
\end{table}

We integrate EVM into the SwinGait framework. As shown in \cref{tab:evm_comparison}, our approach surpasses the performance of EVM. It's noteworthy that EVM operates on extracted features rather than raw images or videos as input. Our method focuses on enhancing feature representation, while EVM enhances the score calculation process.

\subsection{Additional Ablation Experiments}

\begin{table}[h]  % [10]{R}{0.6\textwidth}
    \caption{\small Closed- and open-set face recognition performance.
    [Keys: R@1=rank-1 accuracy; 1\%: FNIR@1\%FPIR]}
    \label{tab:arcface_exp}
    \vspace{-1mm}
    \centering\scriptsize
    \resizebox{0.67\linewidth}{!}{
    \begin{NiceTabular}{lccccc}
        \toprule
        \Block{2-1}{Model} & TinyFace & \Block{1-2}{Surv2Single} & & \Block{1-2}{Surv2Book} & \\ \cmidrule(lr){2-2} \cmidrule(lr){3-4} \cmidrule(lr){5-6}
         & R@1$\uparrow$
 & R@1$\uparrow$
 & 1\%$\downarrow$
 & R@1$\uparrow$
 & 1\%$\downarrow$
 \\  \midrule
        ArcFace & $64.22$ & \(53.73\) & \(68.19\) & \(55.34\) & \(68.37\) \\ \midrule
        ArcFace + IDL & \(64.27\) & \(53.81\) & \(67.23\) & \(55.29\) & \(66.99\) \\
        ArcFace + RTM & \(64.67\) & \(53.84\) & \(67.68\) & \(55.82\) & \(\mathbf{66.16}\) \\ \midrule
        ArcFace + IDL + RTM (Ours) & \(\mathbf{64.70}\) & \(\mathbf{55.26}\) & \(\mathbf{66.70}\) & \(\mathbf{56.51}\) & \({66.86}\) \\
        \bottomrule
    \end{NiceTabular}
    }
\end{table}

We conduct additional ablation experiments using ArcFace with a ResNet-18 backbone on subsets of the TinyFace and IJB-S datasets. 
The results, detailed in \cref{tab:arcface_exp}, validate the effectiveness of both IDL and RTM. These findings showcase our method's adaptability and robust performance across different baselines and biometric scenarios.

\subsection{Analysis on FNIR@FPIR}

\begin{table}[tb]
    \caption{Open-set evaluation of SwinGait-3D and SwinGait-3D with our loss functions. [Keys: Random=random sequences as gallery; Mean=mean feature as gallery]}
    \label{tab:fnir_fpir_analysis}
    \centering\small
    \begin{NiceTabular}{lccc}
    \toprule
        Model & \(R\) & Gallery & FNIR@1\%FPIR \\ \midrule
        \Block{3-1}{SwinGait-3D~\cite{swingait}} & 20 & Random & \(86.56 \smallpm 6.11\) \\
         & 1 & Mean & \(83.50 \smallpm 6.19\) \\ \cmidrule{2-4}
         & 20 & Mean & \(83.50 \smallpm 6.20\) \\ \midrule
        \Block{3-1}{SwinGait-3D~\cite{swingait}+Ours} & 20 & Random & \(82.87 \smallpm 6.72\) \\
         & 1 & Mean & \(79.30 \smallpm 6.71\) \\ \cmidrule{2-4}
         & 20 & Mean & \(79.30 \smallpm 6.74\) \\ \bottomrule
    \end{NiceTabular}
\end{table}

\subsubsection{Effect of \(R\).}
With smaller \(R\), more mated probes become FNs due to failed identification. 
Yet, as shown in \cref{tab:fnir_fpir_analysis}, the hyperparameter \(R\) has a marginal effect on FNIR@1\%FPIR.
The 
This suggests that the bottleneck in open-set biometrics is detection at high thresholds.

\subsubsection{Evaluation Protocol.}
In gait recognition, each subject often possesses multiple gallery sequences. There are different strategies for electing a representative feature from these sequences for each gallery subject.
As discussed in Sec.~\textcolor{red}{4.1}, we take the mean feature for each subject in the gallery in our experiments. To validate this design choice, we compare the open-set FNIR@1\%FPIR on the Gait3D~\cite{gait3d} dataset between the mean gallery feature and the gallery feature of a randomly chosen gait sequence. The results are presented in \cref{tab:fnir_fpir_analysis}. That better results are observed with the mean feature indicates that the mean gallery features are more robust for open-set recognition when available. 

\subsubsection{Effect of \(q\%\).}
\begin{table}[tb]
    \caption{Open-set evaluation of SwinGait-3D~\cite{swingait} and SwinGait-3D with our loss functions. [Keys: Gallery Subj.=number of gallery subjects; NM Probe Subj.=number of non-mated probe subjects]}
    \label{tab:fnir_fpir_analysis_q}
    \centering\small
    \begin{NiceTabular}{ccccc}
    \toprule
        Gallery Subj. & NM Probe Subj. & \(q\%\) & SwinGait & SwinGait+Ours \\ \midrule
        \(140\) & \(35\) & \(20\%\) & \(60.36\smallpm11.83\) & \(57.50\smallpm10.40\) \\
        \(140\) & \(90\) & \(40\%\) & \(69.29\smallpm10.94\) & \(67.50\smallpm11.07\) \\
        \(140\) & \(140\) & \(50\%\) & \(72.73\smallpm9.87\) & \(71.64\smallpm10.41\) \\
        \(140\) & \(240\) & \(60\%\) & \(70.00\smallpm7.99\) & \(68.57\smallpm7.70\) \\
        \(140\) & \(360\) & \(70\%\) & \(69.29\smallpm6.27\) & \(67.14\smallpm6.62\) \\
        \(140\) & \(610\) & \(80\%\) & \(67.50\smallpm4.62\) & \(66.07\smallpm5.43\) \\
         \bottomrule
    \end{NiceTabular}
\end{table}

In evaluating open-set performance, \(q\%\) represents the ratio of probe subjects that are not in the gallery set (Sec.~\textcolor{red}{4.1}). To compare the performance of SwinGait and SwinGait trained with our loss functions under different values of \(q\%\), we create 50 test splits. In each split, we randomly select 140 test subjects as the gallery set and randomly select \num{35}, \num{90}, \num{240}, \num{360}, \num{610} subjects from the remaining subjects as non-mated probes, corresponding to different \(q\%\) values. Since these evaluation protocols use the same set of gallery subjects in each split, they are comparable to each other. 

We present the detailed results of our analysis in \cref{tab:fnir_fpir_analysis_q}. Notably, the gallery set size is relatively small, comprising only 140 subjects, yet the obtained results generally surpass those depicted in Table~\textcolor{red}{4}. Our observations reveal that despite training with only \(p=25\%\), our proposed approach consistently outperforms existing methods across various thresholds of \(q\%\). Moreover, we notice that as the ratio of non-mated subjects increases, the performance of both models initially decreases before showing improvement, indicating that very low \(q\%\) values and very high \(q\%\) values pose more challenging open-set scenarios.

\subsection{Additional Visualizations}

\subsubsection{More Score Distributions.}

\begin{figure}[tb]
    \centering
    
    \begin{subfigure}[b]{0.6\textwidth}
        \centering
        \includegraphics[width=\textwidth]{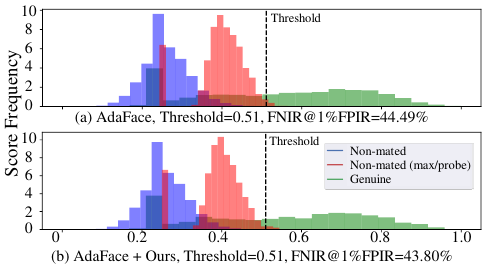}
        \caption{Score distribution on IJB-S~\cite{ijb-s} gallery 1.}
        \label{fig:ijbs_gallery_1}
    \end{subfigure}
    \hfill
    \begin{subfigure}[b]{0.6\textwidth}
        \centering
        \includegraphics[width=\textwidth]{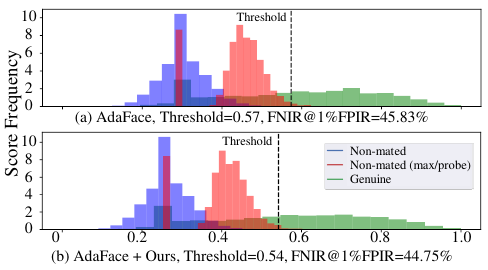}
        \caption{Score distribution on IJB-S~\cite{ijb-s} gallery 2.}
        \label{fig:ijbs_gallery_2}
    \end{subfigure}
    
    \caption{Face recognition and person reID score distributions.}
    \label{fig:face_score_dists}
\end{figure}

We visualize the genuine and non-mated score distributions of AdaFace~\cite{adaface} with and without our loss functions in \cref{fig:face_score_dists}. The official evaluation protocol of IJB-S~\cite{ijb-s} provides gallery 1 and gallery 2 for open-set evaluation. For gallery 1, our loss functions yield a comparable threshold but are able to reduce FNIR@1\%FPIR by improving detection. For gallery 2, our loss functions significantly reduce the threshold, which leads to improved detection as well.

\subsubsection{More FN Analyses.}

\begin{figure}
    \centering
    \begin{subfigure}{0.32\linewidth}
        \includegraphics[width=\linewidth]{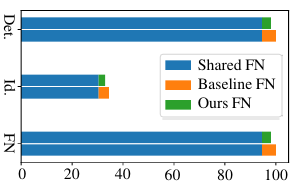}
        \caption{FPIR=$0.1\%$}
        \label{subfig:face_fn_01}
    \end{subfigure}
    \hfill
    \begin{subfigure}{0.32\linewidth}
        \includegraphics[width=\linewidth]{figures/face_fn.pdf}
        \caption{FPIR=$1\%$}
        \label{subfig:face_fn_1}
    \end{subfigure}
    \hfill
    \begin{subfigure}{0.32\linewidth}
        \includegraphics[width=\linewidth]{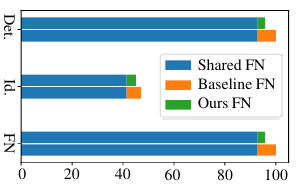}
        \caption{FPIR=$10\%$}
        \label{subfig:face_fn_10}
    \end{subfigure}
    % \hfill
    \begin{subfigure}{0.32\linewidth}
        \includegraphics[width=\linewidth]{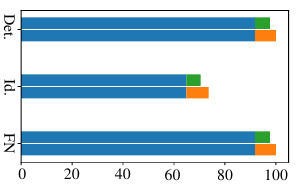}
        \caption{FPIR=$85\%$}
        \label{subfig:face_fn_85}
    \end{subfigure}
    \hfill
    \begin{subfigure}{0.32\linewidth}
        \includegraphics[width=\linewidth]{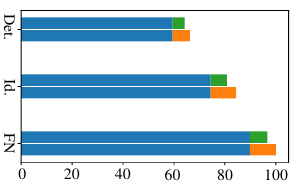}
        \caption{FPIR=$87.5\%$}
        \label{subfig:face_fn_875}
    \end{subfigure}
    \hfill
    \begin{subfigure}{0.32\linewidth}
        \includegraphics[width=\linewidth]{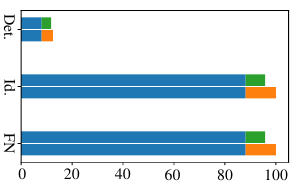}
        \caption{FPIR=$90\%$}
        \label{subfig:face_fn_90}
    \end{subfigure}
    
    \caption{Open-set FNs due to detection, identification, and either. Blue indicates FNs that are shared by the baseline, AdaFace~\cite{adaface}, and our approach. Orange indicates the FNs of the baseline only. Green indicates the FNs of our approach.} 
    \label{fig:fn_fpir_analysis}
\end{figure}

Similar to Fig~\textcolor{red}{6}, we visualize the face recognition FN breakdown on IJB-S~\cite{ijb-s} at different levels of FPIR in \cref{fig:fn_fpir_analysis}. 
At low FPIRs (\cref{subfig:face_fn_01}-\cref{subfig:face_fn_10}), detection is significantly harder than identification as almost all FNs have detection failures and a portion of FNs have both detection and identification failures.

Only at very high FPIR (\eg, \cref{subfig:face_fn_875} and \cref{subfig:face_fn_90}) do a significant portion of FNs have only identification failures. 
The major failure shifts from detection to identification between \(85\%\) FPIR (\cref{subfig:face_fn_85}) and \(90\%\) FPIR (\cref{subfig:face_fn_90}).
However, it is noteworthy that biometrics systems operating at such FPIRs are impractical for real-world use, as they lack the ability to effectively reject unknown subjects.

\subsubsection{Exemplars.}

\begin{figure*}
    \centering
    \includegraphics[width=\linewidth]{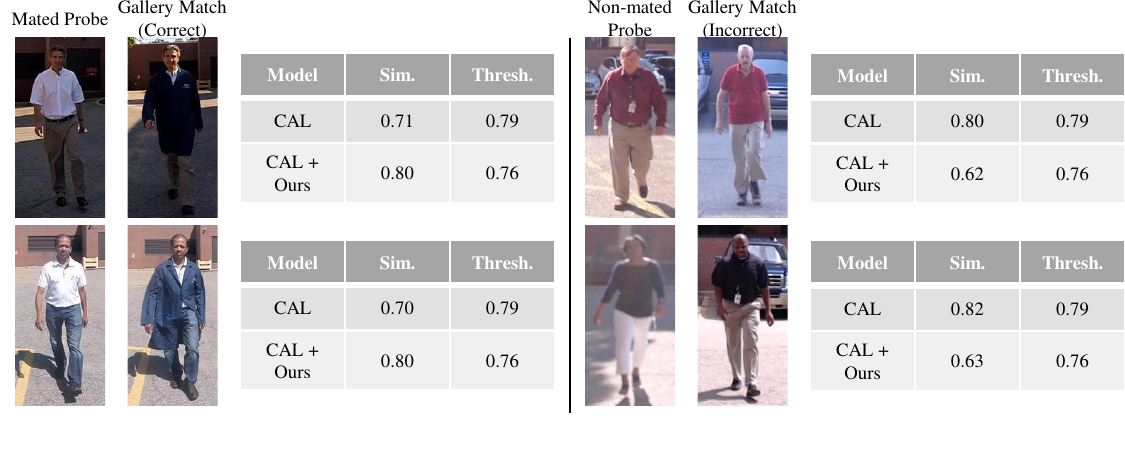}
    \caption{Four probe sequences in CCVID~\cite{cal}. On the left, two mated probes are shown. They are incorrectly rejected by CAL~\cite{cal} but correctly detected with our approach. Two non-mated are shown on the right. They failed to be rejected by CAL, which is corrected by our loss functions.}
    \label{fig:fn_examplars}
\end{figure*}

We visualize the detections of two mated and two non-mated probes in \cref{fig:fn_examplars}. This illustrates that our loss functions can boost both the correct detection of mated subjects and the rejection of non-mated subjects, both of which translate directly to better open-set performance.